\documentclass[twocolumn]{article}

\usepackage{graphicx}      
\usepackage{natbib}        

\usepackage{amsmath}
\usepackage{amssymb}
\usepackage{color}
\usepackage{svg}
\usepackage{graphicx}
\usepackage{enumerate,pgfplots}
\usepackage{tikz}
\usetikzlibrary{arrows,automata}
\usepackage{url}
\title{Survival and Neural Models for Private Equity Exit Prediction\footnote{This work is partially supported by Eurostep Digital AS.}}
\author{Giuseppe C. Calafiore{\small $~^{\#}$}, Marisa H. Morales{\small $~^{\#\#}$}, Vittorio Tiozzo{\small $~^{\#\#}$},\\ Giulia Fracastoro{\small $~^{\#\#\#}$}, Serge Marquie{\small $~^{\#\#\#\#}$}
\vspace{1.6mm}\\
\fontsize{10}{10}
$^{\#}$\,DET Politecnico di Torino and IEIIT CNR\\
$^{\#\#}$\,Politecnico di Torino\\
$^{\#\#\#}$\,DET Politecnico di Torino\\
$^{\#\#\#\#}$\,Eurostep Digital AS\\
\fontsize{9}{9}\selectfont\ttfamily\upshape}
\date{}
\begin{document}
\maketitle



\begin{abstract}                
Within the Private Equity (PE) market, the event of a private company undertaking an Initial Public Offering (IPO) 
is usually a very high-return one for the investors in the company. For this reason,  
an effective predictive model for the IPO event is considered as a valuable tool in the PE 
market, an endeavor in which publicly available quantitative information is generally scarce.
 In this paper, we describe a 
 data-analytic  procedure for predicting  the probability with which  a company will  go public in a given forward period of time. 
 The proposed method is based on the interplay of a neural network (NN) model 
 for estimating the overall  event probability, and 
 Survival Analysis (SA)
 for further modeling 
   the probability of the  IPO event in any given interval of time. 
The proposed neuro-survival model is tuned and tested across nine industrial sectors 
using real data from  the Thomson Reuters Eikon PE database.
\end{abstract}



\section{Introduction}
\label{sec:Intro}
In the last five years, the Private Equity (PE) market saw a sharp expansion of its total economic value due to an unprecedent increase of PE investments  \citep{bainPEreport}. 
However, predicting the outcome of these investments is a very challenging task, since 
the future status of a private company can be extremely uncertain and it can range from bankruptcy, which corresponds to a total loss of the initial investment, to a Initial Public Offering (IPO), the most rewarding event for a PE investor. An impressive example is the investment of Peter Thiel in 2004: \$500,000 which brought him a 693,3\% return by the time of Facebook IPO \citep{facebook_worth}. 
In addition to this extreme volatility of the investment outcome, PE investors have to struggle also with a severe lack of reliable and publicly available information related to the company 
organizational and financial health.
For these reasons, PE fund managers and investors still strive to build  effective strategies
for  identifying private companies that will provide the highest investment returns. 
Currently, most investors rely on naive methods, such as portfolio diversification, to reduce the investment risk. 

In this paper, we tackle the PE exit prediction issue by proposing a data-analytic procedure that estimates the probability of a company to undertake an IPO in a given interval of time. The proposed method is based on an interplay of two models: a neural network (NN) that estimates the overall probability of the event of interest, and a Survival Model (SM) that more finely models this probability as a function of the forward time period. 
Neural networks have been already used in the financial context, see, e.g.,  \citep{fadlalla2001analysis}, in particular they have shown to provide quite strong performance in exctracting valuable information from quantitative data \citep{kim2003discovery,samkin2008adding}.
Instead, Survival Models are broadly employed in the medical field \citep{james2013introduction,swan1996abstinence}, in particular for studying the probability of response to medical treatments in humans and animals, the development of diseases, and the identification and evaluation of prognostic factors and risks related to specific diseases, see \citep{elandt1980survival}. 
SMs have been exploited also in many fields far from the medical ones, such as criminology, industrial production, insurance, and economics, see  \citep{james2013introduction,pitacco2004survival,modarres2016reliability}. To the best of the authors' knowledge, there is however no previous application of SMs to the PE market analysis; in particular,
among the survival approaches, the Accelerated Failure Time (AFT) model \citep{wei1992accelerated}
proved to be the most compatible with the PE framework of interest in this paper.
In a previous work \cite{calafiore2019classifiers} the authors explored a {\em static} model for a private company's exit prediction.
A time-to-IPO analysis, rather than a static analysis, adds an important tool to the PE firm toolbox, in as far as it allows to time appropriately potential intervention. Simply knowing if a potential target will experience an IPO or not, allows a PE firm to decide to invest or not. However, knowing that a company will have an IPO with a given probability in, say, one to two year time, gives to a PE firm a very real opportunity and timing to jump in and invest. 

This paper discusses a time-to-IPO probabilistic analysis for PE firms. The experimental approach that we employed  is summarized as follows:
\begin{enumerate}
\item We considered a Thomson Reuters database containing relevant information  for a large number of companies, see Section~\ref{sec:data} for further details on the database and the features used for constructing the model.
\item Companies are labelled with 4 possible statuses: Bankrupt, Acquisition, IPO, Private. These classes synthesise the four main relevant  outcomes of a company in the PE market.
\item A neural network is trained to estimate the probability for each company of the union event of going bankrupt or being acquired. 
\item We process the dataset by excluding bankruptcies and acquired companies. The dataset filtered in this way contains only companies that went IPO during the observation period or that are still private by the end of it, and constitutes our conditional dataset. 
\item An Accelerated Failure Time (AFT) model is trained on the conditional dataset in order to model, for each company, the probability of occurrence in time of an IPO event. Since we fitted the AFT models on a dataset without bankruptcies and acquisitions, the probability obtained is conditioned on the knowledge that the company has not gone bankrupt or been acquired.
\item The total probability of going IPO in time is eventually computed by de-conditioning the AFT results.
\end{enumerate}

The contribution of this paper is the proposition, description and practical application of a machine learning procedure to the PE field.
We have tested this procedure on a set of real data, and we report
statistically significant results that may  have a high potential impact in this application field.


\section{Input Data}\label{sec:data}
The dataset used in this paper has been extracted from the financial platform Thomson Reuters Eikon \textsuperscript{\textregistered} \citep{eikonPage}. 
This dataset contains the investment data of all the companies founded between 1998 and 2018 in Europe and in the United States. The records with missing data have been excluded. 
After this procedure, we obtained a dataset of 47907 companies belonging to nine business sectors, as shown in Table~\ref{tab:exit_table}.
The available data for each company are: the names of the investors of the first 3 rounds of investment, the foundation date and the IPO date (if the company went public), the number of firms in each round, and the VIX index \citep{vixPage} value on the round dates, which is a public market sentiment indicator that measures the volatility (i.e., the uncertainty) of the public financial market.
Moreover, each company record is labeled with its industrial sector and the actual company status, a categorical value that has been aggregated in 4 labels: IPO, Bankrupt, Acquisition and Private.
The qualitative information of the investor names has been transformed into  quantitative data by means of an investor ranking approach: each investing firm is assigned to a ranking according to the number of rounds in which it has participated. The motivation behind investor ranking is simple: the largest the number of rounds an investor has performed, the strongest is its position in the PE market. 
This approach has been employed also in other machine learning algorithms within the PE framework, see, e.g., \citep{bhat2011predicting,calafiore2019classifiers}.

\subsection{Descriptive statistics}
\begin{figure}
    \centering
    \includegraphics[width = 1\linewidth]{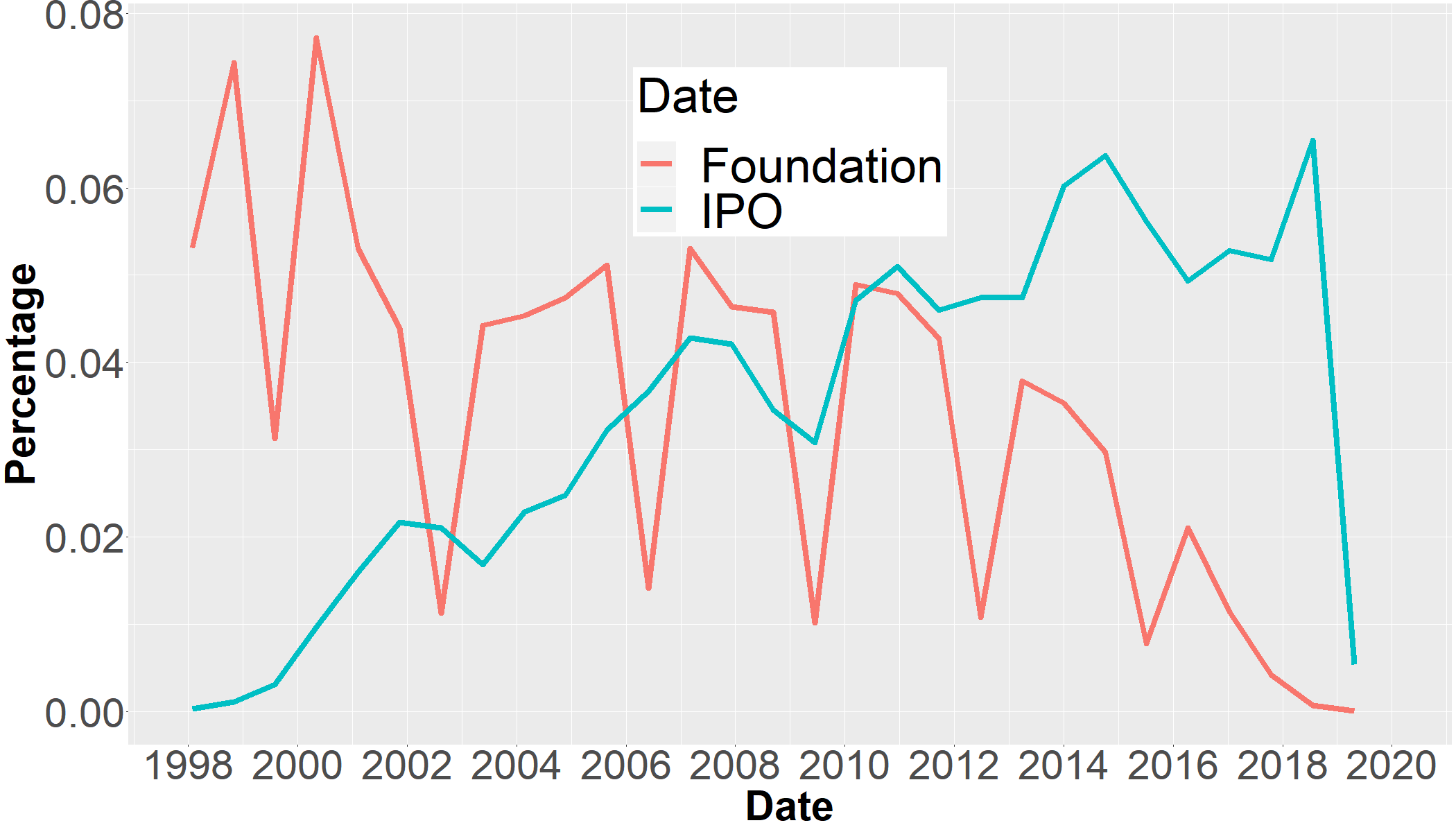}
    \caption{Distribution of foundation (red line) and IPO dates (blue line) of the subjects (companies) under study.}
    \label{fig:Found_IPO_distr}
\end{figure}

Figure~\ref{fig:Found_IPO_distr} shows the time distribution of IPOs and company foundations. It is worth to note how in the recent years the number of company foundations is substantially decreased with respect to the past. On the other hand, the number of IPOs is increased. This behavior shows the maturity of the company population considered in our analysis.

Table~\ref{tab:exit_table} represents the final population distribution across status labels and business sectors. It is clear that our dataset suffers from  class unbalancing, since bankruptcies represents less than $2\%$ of the total data set. This issue has been handled with geometrical resampling techniques, as  described in Section~\ref{sec:NN}.

\begin{table}[]
\centering
\scalebox{0.8}{
\setlength\tabcolsep{3pt} 
    \begin{tabular}{l|r|r|r|r|r|r}
\hline
 Sector & Bank. & IPO & LBO & M\&A & Private & \textbf{Total} \\ 
  \hline
  1 Communications & 165 & 1084 & 3 & 46 & 856 & 2154 \\ 
  2 Computer & 672 & 4926 & 4 & 140 & 11383 & 17125 \\ 
  3 Electronics & 151 & 940 & 1 & 27 & 1734 & 2853 \\ 
  4 Biotech & 84 & 1465 & 3 & 21 & 2704 & 4277 \\ 
  5 Medical & 50 & 1279 & 0 & 13 & 2599 & 3941 \\ 
  6 Energy & 5 & 492 & 1 & 9 & 852 & 1359 \\ 
  7 Consumer & 82 & 1851 & 5 & 33 & 3444 & 5415 \\ 
  8 Industrial & 108 & 2285 & 6 & 34 & 3755 & 6188 \\ 
  9 Other & 96 & 1605 & 4 & 24 & 2866 & 4595 \\
  \hline
  \textbf{Total} & 1413 & 15927 & 27 & 347 & 30193 & \textbf{47907} \\ 
  \hline
    \end{tabular}}
    \caption{Number of companies in each sector for each status within our dataset.}
     \label{tab:exit_table}
\end{table}

\section{Survival Analysis}
\label{sec:SurvCurv}
In this section, we present the basic concepts of Survival Analysis.
Survival Models attempt to estimate the time to a certain event in a probabilistic framework.
In our model, we are interested in estimating for each subject $i$ the ``survival time'' $T_i$, which is defined here as the period of time from the beginning of the subject observation (i.e., its foundation date) to the date $t_{E_i}$ of the event of interest $E_i$ (i.e., the company's IPO date). 
In order to estimate the survival time $T_i$, SM can benefit from the knowledge of additional features regarding the characteristics of each subject. These subject's features will deeply depend on the application domain of the study. In this paper, we consider as additional features four measures of each investment round: the average, the maximum and the minimum ranking of investors who participated in that round, as well as the number of such investors. Moreover, exclusively for the first two rounds the VIX index value is also retained. In total, we consider 14 features for each company.

\subsection{Survival Models Assumptions}
\label{subsec:SM_assumptions}
SA is typically performed either  by means of  non-parametric Cox models \citep{cox1972regression}, or by  the parametric AFT models \citep{wei1992accelerated}.
The first category has the advantage that it does not need an underlying parametric probability distribution. However, Cox models rely on a rather strong  Proportional Hazard Assumption (PHA), which states that the ratio of the hazard functions of two subjects drawn from the population is a constant not depending on time \citep{cox1972regression}.
For each subject $i$, the hazard function $h_i(t)$ is the probability that $E_i$ will happen in the next, very short, time interval $(t,t+\delta t)$ \citep{lee2003statistical_haz}: 
    \[
        h_i(t) = \lim_{\Delta t \to 0} \frac{ \mathbb{P}(t_{E_i}\  \in (t,t+\Delta t)|t_{E_i} \geq t) } {\Delta t}. 
    \]
The PHA implies that
\[
   h_i(t) = h_0(t)g(\Vec{x_i}),
\]
where $\Vec{x_i}$ is the feature vector of subject $i$, $h_0(t)$ is a baseline hazard function that does not depend on the subject, and $g(\cdot)$ is a deterministic function of the subject features. The hazard function ratio of two companies is then
\[
   \frac{h_i(t)}{h_j(t)} = \frac{h_0(t)g(\Vec{x_i})}{h_0(t)g(\Vec{x_j})} = \frac{g(\Vec{x_i})}{g(\Vec{x_j})},
\]
where $\Vec{x_i}$ and $\Vec{x_j}$ are the feature vectors of  subjects $i,j$. 
In the specific application considered in this paper this assumption is quite strong, requiring that the ratio of the probability of going IPO of two companies would depend just on the investors who invested in them, and not on time. 

In order to verify if the considered dataset satisfies the PHA, we computed the Schoenfeld Residuals (SR) \citep{schoenfeld1982partial} for each covariate  and performed the statistical test (SR test) proposed by \cite{grambsch1994proportional}. 
This test is based on the Schoenfeld residuals of a fitted Cox model weighted by its estimated covariance matrix. The formulation of the Schoenfeld residuals is not trivial, as it relies on the first derivative of the Cox regression log-likelihood function, and we refer the reader to \cite{schoenfeld1982partial} for additional details. Using the Schoenfeld Residuals, the PHA becomes equivalent to the hypothesis that the correlation between time and these residuals is null for all the covariates. Therefore, rejecting this hypothesis, even if only for one variable, equals to rejecting the possibility of using Cox models themselves.

The SR test has been performed on a
 business sector by sector basis. 
In Table~\ref{tab:srtest_results} are reported the $p$-values of the SR test computed on all the covariates (Global $p$-values). 
 For each sector, the Global $p$-value is well below $5\%$, therefore the PHA hypothesis can be globally rejected with a $95\%$ significance.
\begin{table*}
\centering
\scalebox{0.95}{
\setlength\tabcolsep{4.5pt} 
    \begin{tabular}{c|c|c|c|c|c|c|c|c|c}
    \hline
 Sector & 1 & 2 & 3 & 4 & 5 &6 & 7 & 8 & 9 \\ 
  \hline
  Global $p$-value &1.21e-12 & 7.36e-72 & 3.88e-14 & 4.37e-08 & 1.20e-06& 4.76e-02 & 1.015e-11 & 5.47e-16 & 2.19e-07 \\ 
  \hline
    \end{tabular}}
    \caption{Results of the Schoenfeld residuals test on the Cox model as a whole for each sector.}
    \label{tab:srtest_results}
\end{table*}

\subsection{AFT model}
Since Cox model assumptions appeared to be invalidated by our data, we considered AFT models, which no not rely on such assumptions.
Within the AFT framework, each $T_i$ takes the form of a parametric random variable whose distribution is assumed to be known (e.g., Exponential, Weibull, lognormal etc.), while its parameters depend on the feature vector $\Vec{x_i}$, which is assumed to be directly related to the survival time $T_i$.

The AFT model uses a log-linear relation to describe the dependency between $T_i$ and the feature vector of the subject, see \citep{lee2003statistical_aft_theory}:
\begin{equation}
    \log T_i = a_0 + \sum^m_{j=1}a_{ij} x_{j} + \sigma\epsilon = \eta_i + \sigma\epsilon, 
    \label{eq:aft_regr}
\end{equation}
where $x_{j}$, $a_{ij}$, and $a_0$ are respectively the $m$ covariates, the $m$ regression coefficients and the intercept parameter. All of these quantities can be summarized with $\eta_i$, which represents the deterministic term. On the other hand, $\epsilon$ is a stochastic error term, equipped with a density function $g(\epsilon)$, scaled by the unknown parameter $\sigma$. The error term $\epsilon$ is the actual source of randomness of the model and its probability distribution determines the resulting $T_i$ distribution. 
It is important to underline that the regression coefficients and the parameter $\sigma$ are fundamental to compute the distribution parameters of each random variable $T_i$. 
As an example, if $\epsilon$ has a standard normal distribution, then each $T_i$ will have a lognormal distribution having mean $\mu_i = a_0 + \sum^m_{j=1}a_{ij} x_{ij} = \eta_i$ and standard deviation $\sigma_i = \sigma$.
We compute the regression coefficients and the parameter $\sigma$ employing a Maximum Likelihood Estimation (MLE). 
To test the statistical significance of the covariates, we used the Wald's statistic \citep{huber19721972}, whose results are reported in Section~\ref{sec:results}. 

\section{Model fitting} \label{sec:ExpDesign}
In this section, we describe the procedure employed in order to fit the proposed survival model to the PE data.

\subsection{Data preprocessing}
When we apply SMs to PE data, the first problem encountered is the presence in the dataset of bankrupt and acquired companies. Indeed, plain SMs are able to model only one event and do not consider the case when the possible outcomes are more than one. More critically, 
the timing information about bankrupt or acquisition events is rarely available in the database.
To deal with such multiple outcomes, we preprocess the dataset excluding the companies that went bankrupt or have been acquired. Then, we build the AFT model on the remaining observations (IPO and Private). This means that the probability that we estimate is conditioned by the fact that we know that the company has not experienced bankrupt or acquisition during the observation period. In addition, we divide the dataset according to the companies' industrial sector, and we train a different SM for each sector. 
 
\subsection{Censoring}
A typical  problem that we face when we apply SMs to PE data is the presence of companies whose status remains  private within the data time span. 
These companies enter the observation period at certain times (the foundation date) and by the end of the study the IPO event has not occurred yet. 
We computed their survival times as the difference between the end of the observation period and their foundation date, and we marked with a binary variable these companies as censored observations. 
The capability of dealing with censored data is actually a key property of Survival Models. 
The censoring label encodes 
the fact that for a censored company we do not know the exact survival time, but nevertheless we know that this survival time is {\em larger} than the observation time, and this information is exploited in the computation of the survival probability, see, e.g., \citep{lee2003statistical_censoring}. 

\subsection{AFT procedure} \label{subsec:aft_procedure}
We use the AFT model to estimate the conditional survival probability $\mathbb{P}(T_i>t|V_i)$, where $V_i$ represents the event that the company $i$ does not experience bankrupt or acquisition during the observation period. 
It is worth mentioning that in the analysis of the PE market the quantity of interest is the probability for a company $i$ of going IPO in a given time interval (i.e., $\mathbb{P}(T_i\leq t)$). However, since SMs are mainly used in the biomedical field, they are designed to model the survival probability $\mathbb{P}(T_i>t)$. Clearly, 
it holds that 
$\mathbb{P}(T_i\leq t) = 1 - \mathbb{P}(T_i>t)$. 

In order to capture the different sector characteristics, we estimated a different SM for each of the nine industrial sectors. To train the nine AFT models, for each sector we have to made two choices: which distribution to choose for the stochastic error term $\epsilon$ and which covariate to use. We decided to test 4 different distributions in order to explore 3 different families: Exponential (Exponential, Weibull), Normal (LogNormal) and Snedecor's F (Generalized F) \citep{cox2008generalized}. 

\begin{figure*}[tb]
    \centering
    \includegraphics[width = 1\textwidth]{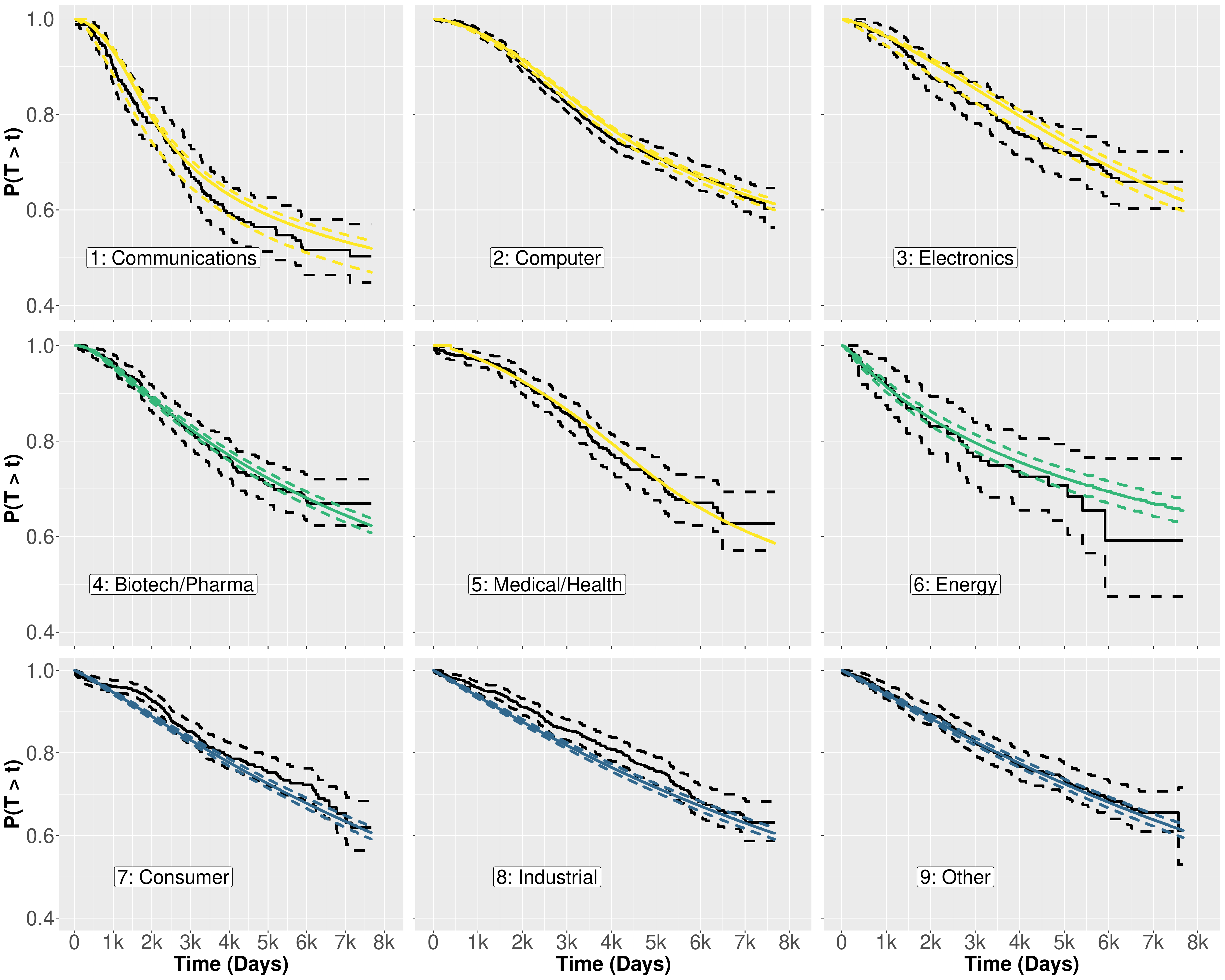}
    \caption{Unconditioned AFT fitting compared to KM estimate built on the validation set, sector by sector. Black lines: Kaplan-Meyer empirical survival estimate of the validation set. Colored lines: best AFT fitted distributions on fitting test, for each sector. Weibull (blue line), Lognormal (green line), and Generalized F (yellow line) distribution. Dashed lines: $95\%$ confidence intervals.}
    \label{fig:aft_val_uncond}
\end{figure*}

Our model selection procedure relies on a numerical (Likelihood, p-value) 
%
and graphical (fitting plots) evaluation, which has been performed on a training dataset. First, for each distribution $k$ and for each sector $s$ a separate model $M_{k,s}$ has been fitted and the statistical significance of the covariates has been verified with the Wald test. 
%
Then, each $M_{k,s}$ has been fitted again on the $90\%$ significant variables found previously for that model itself and its Maximum Log-Likelihood Estimate (MaxLLE) recorded. 
%
Finally, the unconditioned AFT survival fitting curves have been plotted against the validation set curves (Figure~\ref{fig:aft_val_uncond}), that were obtained through a Kaplan-Meier \citep{kaplan1958nonparametric} empirical estimate (KM curve). The latter computation relies on a simple and robust counting process that is widely recognized as a consistent reference to benchmark graphical accuracy of the AFT fitted models \citep{lee2003statistical}. The results of this model selection procedure are discussed in Section~ \ref{sec:results}.

After having estimated the conditional probability $\mathbb{P}(T_i>t|V_i)$ with the procedure described above, we compute the marginal probability $\mathbb{P}(T_i\leq t)$. 
By the law of total probability, the following holds:
\begin{equation}
\begin{split}
\mathbb{P}(T_i \leq t) &=1-\mathbb{P}(T_i > t) \\
&= 1- \mathbb{P}(T_i > t |V_i) 
\mathbb{P}(V_i),
\end{split}
\label{eq:marg_prob}
\end{equation}
where $\mathbb{P}(V_i)$ has been estimated, for each company, using the neural network described in the following section. The second equality holds because we have assumed that $\mathbb{P}(T_i > t |\bar{V}_i)$=0, where $\bar{V}_i$ is the event that the $i$-th company has been acquired or has gone  bankrupt during the observation period. This seems a fairly reasonable assumption, since if a company was acquired or went bankrupt, it cannot go IPO, and the case where a public company go bankrupt is so rare that it is not worth taking account of it.

\subsection{Neural estimation}
\label{sec:NN}
In this section we describe the simple neural network (NN) approach that 
was used to estimate the probability $\mathbb{P}(\bar{V}_i)$ that the $i$-th company is acquired or goes bankrupt. The input data for the NN are the features described in Section~\ref{sec:data} except for the Foundation date and the IPO date. Neural Networks have proven to achieve strong predictive performance within the financial framework \citep{fadlalla2001analysis}, especially for interpreting patterns from qualitative data \citep{kim2003discovery,samkin2008adding}. Moreover, it has already been shown that the selected features can constitute a strong basis for Machine Learning algorithms within the Private Equity framework \citep{bhat2011predicting}.    

The proposed neural network is a 1-layer Multilayer Perceptron (MLP). We use as activation function the Scaled Exponential Unit (SeLU) \citep{klambauer2017self}, since it provides strong evidences in preventing gradient vanishing optimization issues and promotes information flow through the net. Each company has been labeled as Bankrupt-Acquisition (BA) or Non-Bankrupt-Acquisition and the global dataset has been split into a training and a test set with a $2/3$--$1/3$ ratio.
Since the training dataset is  unbalanced  (the BA companies are 1787 out of 47907), a geometrical resampling technique has been used, namely the SVMSMOTE described in \citep{tang2008svms}. 
The numerical optimization procedure has been performed using the Adam algorithm \citep{kingma2014adam}, a well-know method with self-adapting moment estimation, and the network has been trained for 70 epochs. 

The performance metrics for both classes are shown in Table~\ref{tab:nn_results}. The Precision measures the ratio of elements correctly identified over the total elements predicted in that class, the Recall measures the ratio of elements actually identified over the total number of elements present in that class in the validation set, while the accuracy represents the overall precision measure in both classes. The output probability of the net was classified as {BA} if it was over $50\%$. The results obtained are comparable to the ones found in literature for a Machine Learning classifier on a Private Equity dataset \citep{bhat2011predicting,calafiore2019classifiers}, even if these related works act on a more balanced dataset.

\begin{table}[tb]
        \centering
        \scalebox{0.89}{
        \setlength\tabcolsep{4.5pt} 
        \begin{tabular}{l|l|l|l|l}
        \hline
            Precision+ & Recall+ &  Precision- & Recall- & Accuracy\\
            \hline
            0.10 & 0.99 & 0.67 & 0.81 & 0.81  \\
            \hline
        \end{tabular}}
        \caption{NN predictive performance: ``+'' refers to BA event, ``-'' to Not-BA. 
        }
        \label{tab:nn_results}
\end{table}

\section{Results}\label{sec:results}
In this section, we present the experimental results of the neuro-survival model described in the previous sections.
In order to fairly evaluate the proposed models, we divide the original dataset into a training and a validation set according to a $9/10$--$1/10$ proportion.

As discussed in Section~\ref{subsec:aft_procedure}, for each model $M_{k,s}$ we have to choose the distribution of $\epsilon$ and select the significant covariates. The Wald's test for each $M_{k,s}$ resulted in a number of significant covariates reported in 
Table~\ref{tab:num_sel_covs_aft_completa}. 
Instead, the distribution choice is the result of an integrated approach. It has to maximize the MaxLLE, minimize the number of Wald's selected covariates, and well approximate the KM curves. 
As for the MaxLLE absolute value for the models fitted with the selected covariates, the difference across distributions in each sector was extremely low compared to its order of magnitude. So, this metric alone could not be considered as a robust approach for selecting the best fitting distribution. However, the distributions with the highest MaxLLE are the ones with the least number of selected variables in the Wald's test, in all sectors except sector 9. Then, in order to definitively select the distribution, we have analysed the fitting plots choosing the distributions  whose fitting curves  better capture the KM curves pattern. Table~\ref{tab:num_sel_covs_aft} reports the distribution and the number of covariates chosen for each sector. 
The three criteria agree on the best distribution for all sectors except sectors 4 and 9. 
In sector 4, the LogNormal distribution was found to approximate the KM curve better than the Genaralized F, which was the distribution with the highest MaxLLE. This was evident on the right tail of the KM curve, where the Genaralized F lied outside the KM confidence intervals, for this reason the LogNormal distribution was preferred.
In sector 9, according to the Wald's test, Exponential or LogNormal distribution were the best choices, however the Weibull distribution showed a better fitting and a higher MaxLLE. This partial contradiction can be related to the complex empirical pattern of this sector, since it groups all the companies without a specific industrial activity classification.

We tested the models obtained with this procedure on the validation dataset and computed their marginal probability $\mathbb{P}(T_i \leq   t)$ using Eq. \eqref{eq:marg_prob}. Figure~\ref{fig:aft_val_uncond} shows the fitting curves compared to the KM estimate. We can see that the obtained curves are strongly consistent with the validation data. In each sector, the selected models (colored lines) fit the empirical estimates (black lines)  reliably: fitting curves are within the KM estimates confidence interval, except for a few very restricted areas. 

\begin{table}[tb]
\begin{center}
\begin{tabular}{l|rrrr}
  \hline
Sector & Exp. & Wei. & LogNorm. & Gen. F \\ 
  \hline
1 & 8 & 10 & 8 & 6 \\ 
  2 & 10 & 9 & 8 & 7 \\ 
  3 & 7 & 7 & 7 & 7 \\ 
  4 & 8 & 8 & 6 & 6 \\ 
  5 & 3 & 3 & 3 & 3 \\ 
  6 & 0 & 0 & 1 & 0 \\ 
  7 & 2 & 2 & 2 & 0 \\ 
  8 & 3 & 3 & 3 & 0 \\ 
  9 & 2 & 3 & 2 & 3 \\ 
   \hline
\end{tabular}
 \caption{Number of Wald's test $90\%$ significant covariates for each sector and distribution}
 \label{tab:num_sel_covs_aft_completa}
\end{center}
\end{table}

\begin{table*}[t]
\begin{center}
 \scalebox{0.83}{
 \setlength\tabcolsep{3pt} 
\begin{tabular}{l|r|r|r|r|r|r|r|r|r}
  \hline
Sector (selected distr.) & 1 (Gen.F) & 2 (Gen.F) & 3 (Gen.F) & 4 (LogNorm.) & 5 (Gen.F) &6 (LogNorm.) & 7 (Wei.) & 8  (Wei.) & 9  (Wei.) \\
  \hline
\# sign. covariates &  6 & 7 & 7 & 6 & 3 & 1 & 2 & 3 & 3\\
   \hline
\end{tabular}}
\caption{Number of covariates with a statistical significance over 90\% and highest MaxLLE in that sector.} 
\label{tab:num_sel_covs_aft}
\end{center}
\end{table*}

\section{Conclusions}
\label{sec:Conclusions}
In this paper, we presented a predictive model that estimates the time-to-IPO probability of private companies. The proposed method is based on an interplay of a neural network and a AFT survival model.
The results show that such neuro-survival model in the PE framework is able to reliably represent the probability in which a private company will go public in a given time interval. 

The proposed method can be a useful support tool for PE investment decision and {\em timing}, enabling investors to estimate the probability of a company to experience a desirable IPO event within a given time interval or,
by complement, the probability that a PE investment will result in an adversary event.
%


\bibliography{ifacbib.bib}             

\end{document}